\documentclass{article}

\usepackage[english]{babel}
\usepackage{amssymb}
\usepackage{mathtools}
\usepackage{bbm}
\usepackage{amsmath}
\usepackage[letterpaper,top=2cm,bottom=2cm,left=3cm,right=3cm,marginparwidth=1.75cm]{geometry}

\usepackage{graphicx}
\usepackage{natbib}
\usepackage[colorlinks=true, allcolors=blue]{hyperref}

\newcommand{\tr}{\top}
\newcommand{\EE}{\mathbbm{E}}
\title{Baird Counterexample is Solved:\\
with an example of How to Debug a Two-time-scale Algorithm}
\author{Hengshuai Yao}

\begin{document}
\maketitle

\begin{abstract}
Baird counterexample was proposed by Leemon Baird in 1995, first used to show that the Temporal Difference (TD(0)) algorithm diverges on this example. Since then, it is often used to test off-policy learning algorithms. Gradient TD algorithms solved the divergence issue of TD on Baird counterexample. However, their convergence on this example is still very slow, and the nature of the slowness is not well understood to date, e.g., see \citep{sutton2018reinforcement}. 

This note is to understand in particular, why TDC is slow on this example, and provide a debugging analysis to understand this behavior. Our debugging technique can be used to study the convergence behavior of two-time-scale stochastic approximation algorithms. 
We also provide empirical results of the recent Impression GTD algorithm on this example, showing the convergence is very fast, in fact, in a linear rate. We conclude that Baird counterexample is solved, by an algorithm with the convergence guarantee to the TD solution in general, and a fast convergence rate.   
\end{abstract}

\section{Introduction}
\begin{figure}[h]
\centering
\includegraphics[width=0.7\linewidth]{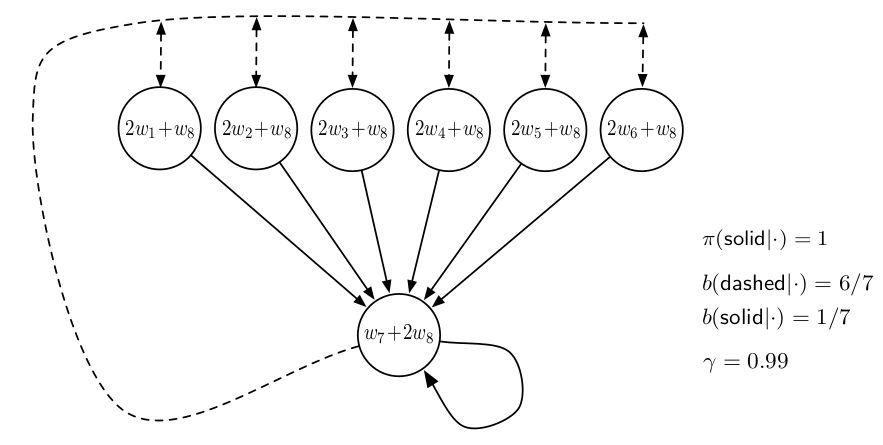}
\caption{Baird Counterexample. Image is from \citep{sutton2018reinforcement}.}
\label{fig:baird}
\end{figure}

This short note assumes that  the readers have a good knowledge of off-policy learning. For relevant background, one can refer to Chapter 11 of \citep{sutton2018reinforcement}. This note arises from a debugging experimentation for off-policy learning on Baird counterexample (Figure \ref{fig:baird}) \citep{baird1995residual}. Part of the results are in \citep{impressionGTD}. Due to space limitations, most of this experiment wasn't presented there. For Baird counterexample is an important problem in the history of off-policy learning in reinforcement learning, we think it is worth updating the literature on the latest status of solving this problem. The technique we used for debugging TDC is also useful in the future if someone would like to understand the behavior of two-time-scale stochastic approximation algorithms.      

\begin{figure}
\centering
\includegraphics[width=0.6\linewidth]{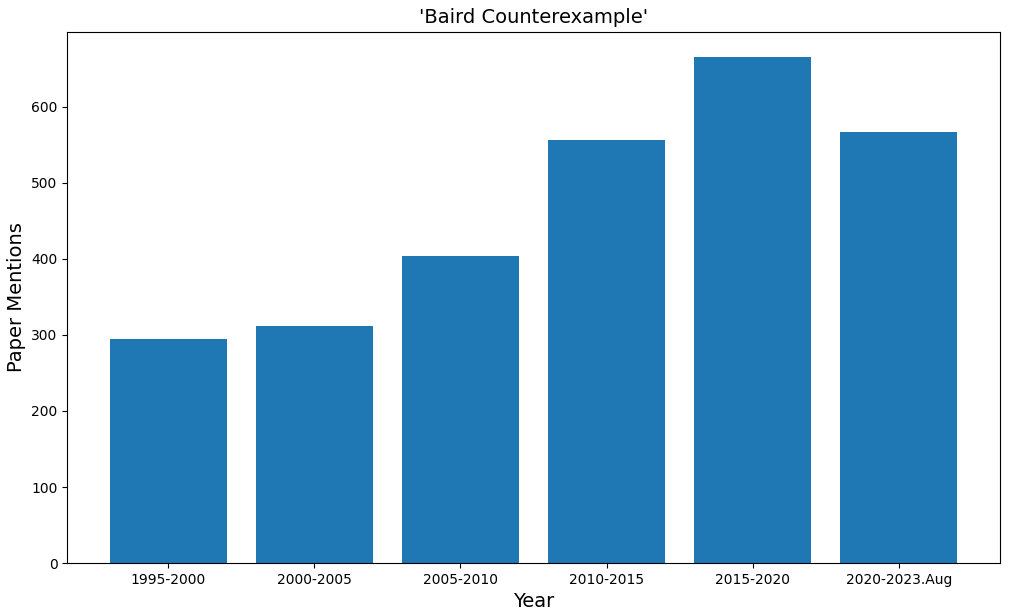}
\caption{Number of papers that mentioned ``Baird Counterexample'' over the years. Data was retrieved using Google scholar.}
\label{fig:papers}
\end{figure}

Figure \ref{fig:papers} shows the number of papers that mentioned ``Baird Counterexample'' over the years. Until August 2023, there are about 3,000 papers that mentioned this example.  
Many reinforcement learning papers used this example as a test bed for their off-policy learning algorithms.

In Figure \ref{fig:baird}, Baird counterexample is a Markov Decision Process problem, in which there are transitions between state nodes, indicated by the arrows. There are seven states. Two policies that specify how they would select the arrow. There are two kinds of arrows, dashed and solid. A policy $\pi$ selects the solid arrow with probability one. Another policy $b$ selects the solid arrow with probability $1/7$, and the dashed arrow with probability $6/7$. At the state below, after selecting the dashed arrow, we go to the six upper states with an equal probability of $1/6$. 
All the rewards are zero. Thus the value function for each state is zero. 

This problem is an off-policy learning problem. An learning agent follows policy $b$ to surf among the seven states, but its goal is to learn the value function of $\pi$. The value function is parameterized by eight weight parameters, which are to be learned by the agent. Clearly, the features are powerful enough for representing the value function exactly. We thus expect off-policy learning algorithms can find the correct value function, which is zero for every state. Let $\gamma$ be the discount factor in the value function. In our experiment, we used $\gamma =0.9$.

\section{The TDC Algorithm}
We focus on the TDC algorithm \citep{tdc}, which is a gradient TD algorithm.  See \citep{impressionGTD} for a review of the algorithm. 

The algorithm processes transitions in the form, $(\phi, \phi', r)$, where $\phi$ is the feature vector of the state where we select an arrow action, and $\phi'$ is the feature vector of the resulting state, and $r$ is the reward.   
TDC is an incremental algorithm that has two updates:
\begin{align}\label{eq:tdc}
    \theta_{t+1} &= \theta_t - \alpha_t \left[- \delta_t(\theta_t)\phi_t + \gamma \phi_t' (\phi_t^\tr  w_t)\right], \nonumber \\
    w_{t+1} &= w_t - \beta_t(\phi_t^\tr  w_t-\delta_t)\phi_t, 
\end{align}
where $\alpha_t, \beta_t$ are the step-size parameters. We call $\theta_t$ the main iterator and $w_t$ the helper iterator.

TDC is a typical two-time-scale update, because the convergence can be guaranteed if $\alpha_t$ and $\beta_t$ are of different orders. In particular, $\beta_t$ is often larger than $\alpha_t$, meaning that $w$ is updated faster than $\theta$. In practice, though, $\alpha_t$ and $\beta_t$ are often set to constants, which is used by \citep{sutton2018reinforcement}. The well-performing $\alpha_t$ and $\beta_t$ combinations often find their optimal values are different.

TDC is related to the MSPBE objective function: 
\[
\bold{MSPBE}(\theta)= (A \theta + b)^\tr  C^{-1} (A \theta + b), 
\]
where $\theta$ is our learning weight vector, and $A, b$ are defined by $A=\EE[\phi(\gamma \phi' - \phi)^\top]$, $b=\EE[\phi r]$, and $C = \EE[\phi \phi^\tr]$. 
The O.D.E. underlying TDC involves the gradient of the MSPBE function.

\section{Performance of TDC}

The best result of TDC for Baird counterexample can be found in \citep{sutton2018reinforcement}, which drives the MSPBE close to zero. However, TDC still cannot find a good estimate of the value function in a reasonable amount of time. The value function error is still large even though the MSPBE is near zero. Figure \ref{fig:baird_tdc}, from \citep{sutton2018reinforcement}, shows that value function error ({\color{green}green line}) does not approach zero in a meaningful rate. Instead, the learning seems to halt or get stuck for some reason, failing to make progress after just hundreds of steps. It was understood as a slow learning rate, although a flat curve, that it still converges to zero in the limit \citep{sutton2018reinforcement}. From a practitioner's view, it looks more like a bias. Is it a slow rate or a learning bias? This is not clear, and the phenomenon has not been understood so far. 

The left plot of the figure also shows that the difference in the values of state 1 and state 6, $V(1)-V(6) = 2(w_1-w_6)$, does not approach zero after a reasonably long time for this small problem.  This is strange, because even though their values are NOT learned correctly (zero), they should be learned to be close, due to that the TD error, which is the sampled Bellman error, should be driven to zero by TDC. However, if we look at the MSPBE ({\color{red}red line}), it shows this error correctly converges to zero at the same time of the value function learning error getting stuck. This phenomenon of ``MSPBE converging to zero fast but not the value function learning error'' is puzzling. We are going to understand this phenomenon in our experiment.     

First, we start with the trickiness of tuning TDC. 

\subsection{Why TDC is Unsatisfactory}
Results on Gradient TD algorithms usually only show TDC with well-performing step-size combinations \citep{tdc}. In practice, how sensitive are they to the choice of the two step-sizes?  

\begin{figure}[t]
\centering
\includegraphics[width=0.7\linewidth]{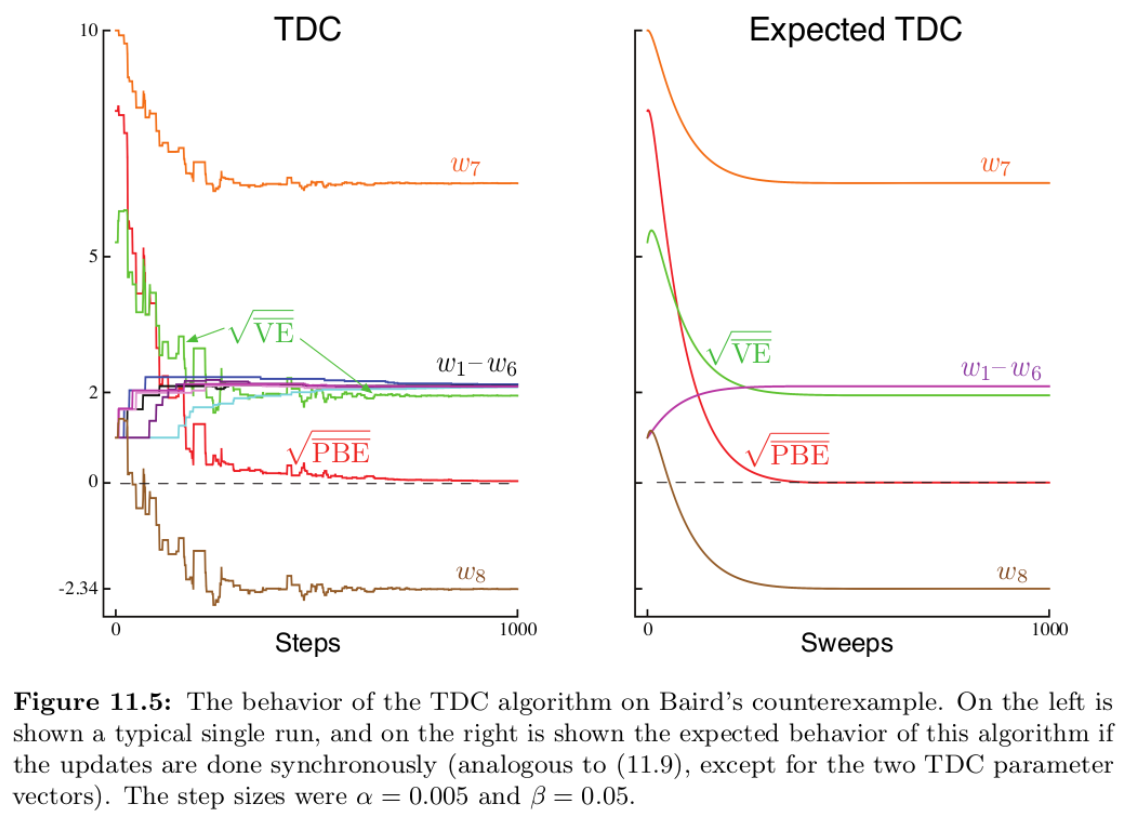}
\caption{TDC on Baird counterexample, from Chapter 11 of \citep{sutton2018reinforcement}.}
\label{fig:baird_tdc}
\end{figure}

\begin{figure}[t]
\centering
\includegraphics[width=0.9\linewidth]{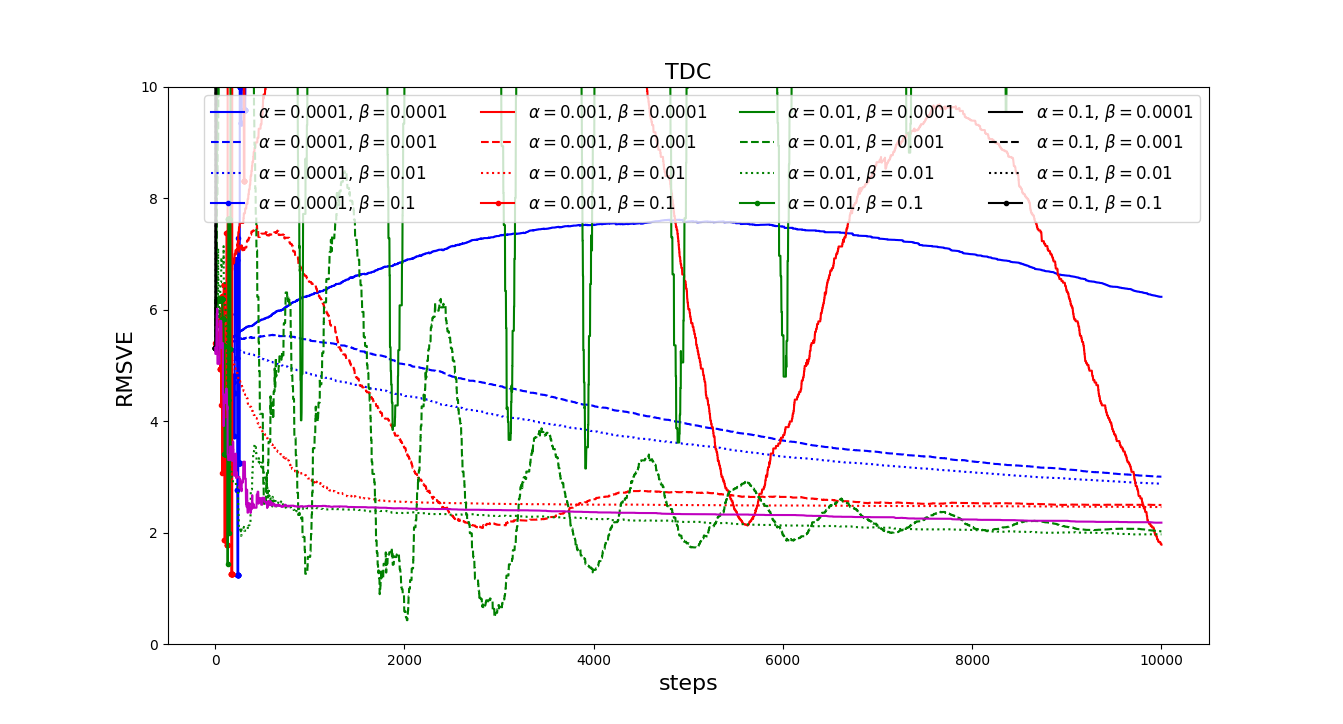}
\caption{TDC performance with different choices of $\alpha$ and $\beta$.}
\label{fig:baird_tdc_alphas_betas}
\end{figure}

TDC with a variety of $\alpha$ and $\beta$ combinations are shown in Figure \ref{fig:baird_tdc_alphas_betas}. The plot cuts to focus on between $y=0$ and $y=10$. Out of this window, the algorithm either diverge or has huge oscillations. The magenta line is 
for $\alpha=0.005$ and $\beta=0.05$, used in the TDC paper and the text book.
Usually, the two-time-scale theory requires that step-size $\alpha$ should be smaller than $\beta$ because the O.D.E. analysis is based on that the first iterator has to wait until the second iterator somehow finishes the averaging or regression job. However, in practice, it is still interesting to see the behavior of the algorithm when $\alpha$ is equal to or bigger than $\beta$. These out-of-theory hyper-parameter choices are sometimes actually used in the experimentation of TDC, e.g., see \citep{martha2020gradient}.  

We see a good set of algorithmic behaviors in Figure \ref{fig:baird_tdc_alphas_betas}. Some choices produce well-behaved learning curves, like the magenta line, which used the step-size choice selected by \citep{tdc}. The blue dotted line is also stable, but very slow. However, the green line with dot marker, although $\alpha$ is also smaller than $\beta$, following the usual principle, has a huge oscillation. Others, mostly have oscillatory behaviors in learning.   

In Figure \ref{fig:baird_tdc_alphas_betas}, we also run much longer learning time than \citet{sutton2018reinforcement}'s experiment in Figure \ref{fig:baird_tdc}. It shows the learning continues to be stuck, and the best of all the error curves is still flat, although some choices produced slightly smaller learning errors than the default choice. None of the hyper-parameter choice leads to a satisfactory solution. The red line hits a low error near the end of the shown time window, but it will quickly go up again as we can predict from the previous learning curvature. This rules out the possibilities that the slow learning of TDC is caused by the hyper-parameter choice, or failure to give it more time of learning. 

\begin{figure}[t]
\centering
\includegraphics[width=0.8\linewidth]{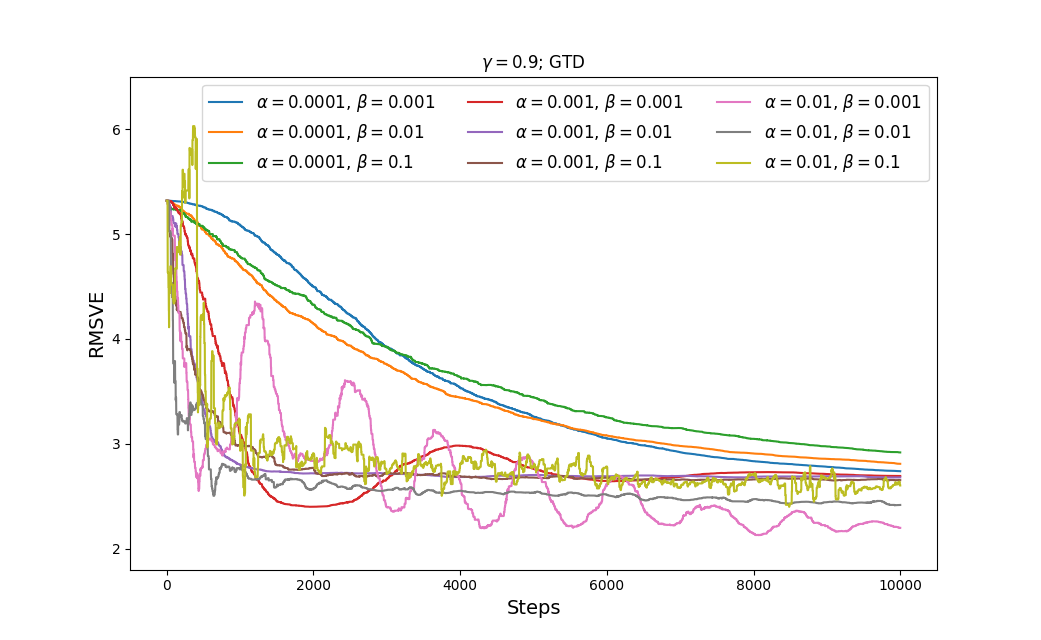}
\caption{GTD performance with different choices of $\alpha$ and $\beta$.}
\label{fig:baird_gtd_alphas_betas}
\end{figure}

We also did experiments of GTD \citep{gtd} for Baird counterexample. The results of using different combinations of the two step-sizes are shown in Figure \ref{fig:baird_gtd_alphas_betas}. The plot below actually shows that the performance of $\alpha$ being equal to or bigger than $\beta$ is not that bad sometimes, e.g., $\alpha=\beta=0.01$.
Comparing to TDC in Figure \ref{fig:baird_tdc_alphas_betas}, we can see that TDC is more sensitive to the choice of the two step-sizes than GTD. Although some choices also lead to oscillation in GTD, the oscillating magnitude 
is much smaller. 

Thus TDC is still not a satisfactory off-policy learning algorithm, because it is very sensitive to the choice of the two step-sizes, besides the slowness that was already illustrated by \citep{sutton2018reinforcement}.

The GTD experiment also reveals an interesting phenomenon.  Using a bigger $\alpha$ learns eventually faster after some steps. However, there seems to be a trade-off between the learning speed and the oscillation magnitude. In particular, to achieve a fast rate, the algorithm has to experience significant oscillation for a long time. This is also true with TDC. The green dashed and dotted lines in Figure \ref{fig:baird_tdc_alphas_betas} both have a smaller learning error than the default choice in the end, for which the $\alpha$ is bigger, i.e., 0.01 versus 0.005. However, both the green lines have significant oscillation. The oscillation with the dashed line lasts a long time. 
Last but not the least, both step-size choices are not covered by the two-time-scale theory (that $\alpha$ should be smaller than $\beta$). Using a small $\alpha$, the oscillation is curbed, but the learning rate is slow (see Figure \ref{fig:baird_tdc_alphas_betas} and also Figure \ref{fig:baird_gtd_alphas_betas}).
This is another baffling phenomenon for two-time-scale Gradient TD algorithms. This probably means that adapting the step-sizes of Gradient TD algorithms is a very hard problem.      

\section{Debugging TDC}
Overall, the best performing step-size combo in Figure \ref{fig:baird_tdc_alphas_betas} is $\alpha=0.01$ and $\beta=0.01$. The value function error (RMSVE) for this best hyper-parameter combo is still above 2.0 after 10,000 steps. The performance of this step-size combo is just slightly better than in the TDC paper, which used $\alpha=0.005$ and $\beta=0.05$. So we simply used their choices in the following experiment. 

We also confirmed that the MSPBE gets reasonably close to zero for TDC in the our experiment.
So what's going on in TDC for the flat learning curves in Figure \ref{fig:baird_tdc} and \ref{fig:baird_tdc_alphas_betas}?

To answer this question, we employ a special experimentation method, that allows us to see it clearer inside TDC. Just monitoring MSPBE and the value function learning error isn't enough to gain a deep understanding of TDC. Our main lens is the regression error by the helper iterator: 
\[
RMSRE(w_t|\theta_t) = \sqrt{\frac{1}{7}\sum_{s=1}^7 (\delta_t(\theta_t) - \phi(s)^\tr w_t)^2}, 
\]
where we calculate the RMS regression error, using the helper iterator $w_t$, given the latest main iterator where the TD errors on the states are realized. 

\begin{figure}[t]
\centering
\includegraphics[width=0.8\linewidth]{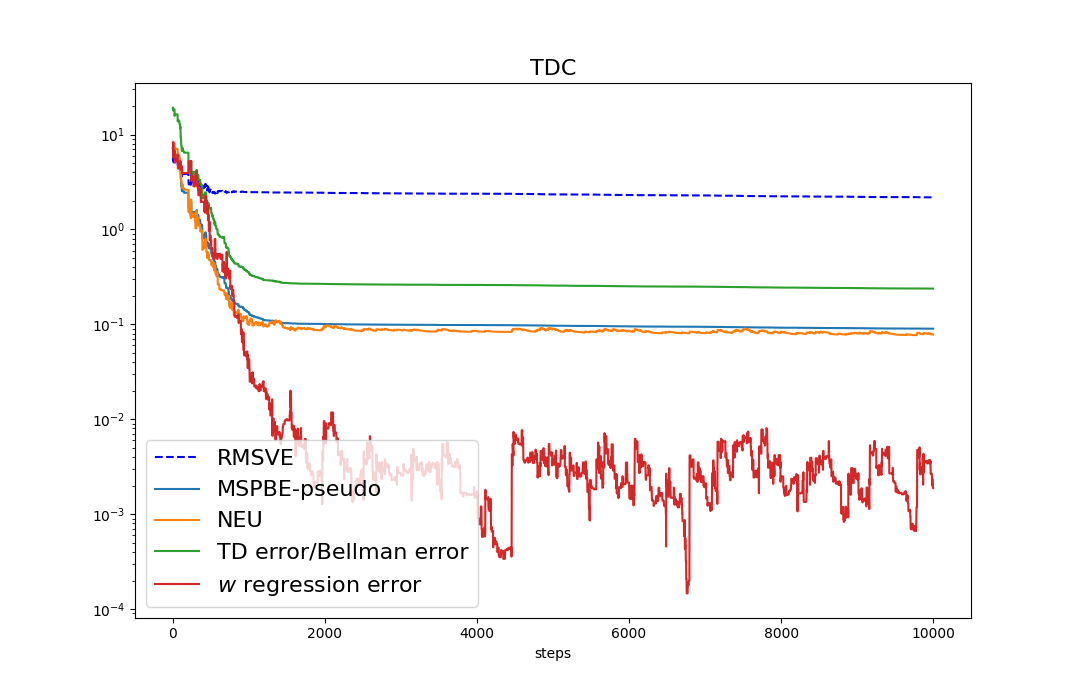}
\caption{Plotting the regression error by the helper iterator.}
\label{fig:baird_w_regression}
\end{figure}

The result of monitoring this loss alongside with others such as NEU and MSPBE are shown in Figure \ref{fig:baird_w_regression}.
Surprisingly, this shows all the other errors shown are high and flat, while the regression error is driven to low quickly. This means in TDC, at least the helper iterator does a good job that it is expected to do: TD error is well predicted, for all states. 

\begin{figure}[t]
\centering
\includegraphics[width=0.8\linewidth]{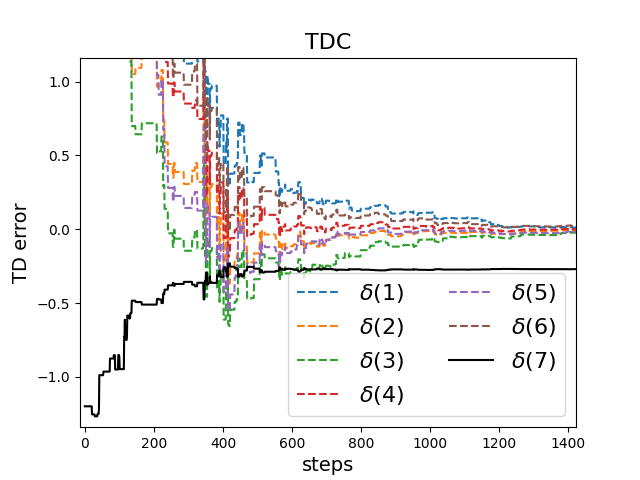}
\caption{TD error profiling for the seven states.}
\label{fig:baird_td_errs}
\end{figure}

However, why the TD error is still large after a long time of learning in Figure \ref{fig:baird_w_regression}? The TD error metric plotted is the weighted TD errors over the states. To see clearly, we profile the TD error under the target policy for each of the seven states, shown in Figure \ref{fig:baird_td_errs}. This shows that all the states’ TD errors converge to zero, except state 7 (the state that is located under all the others in Figure \ref{fig:baird}). State 7 is a crucial state for this example. Its TD error being high means $V(7)$ is not correctly estimated. This in turn causes a big estimation error for the values of all the other six states, though their TD errors are already zero. 

\begin{figure}[t]
\centering
\includegraphics[width=0.8\linewidth]{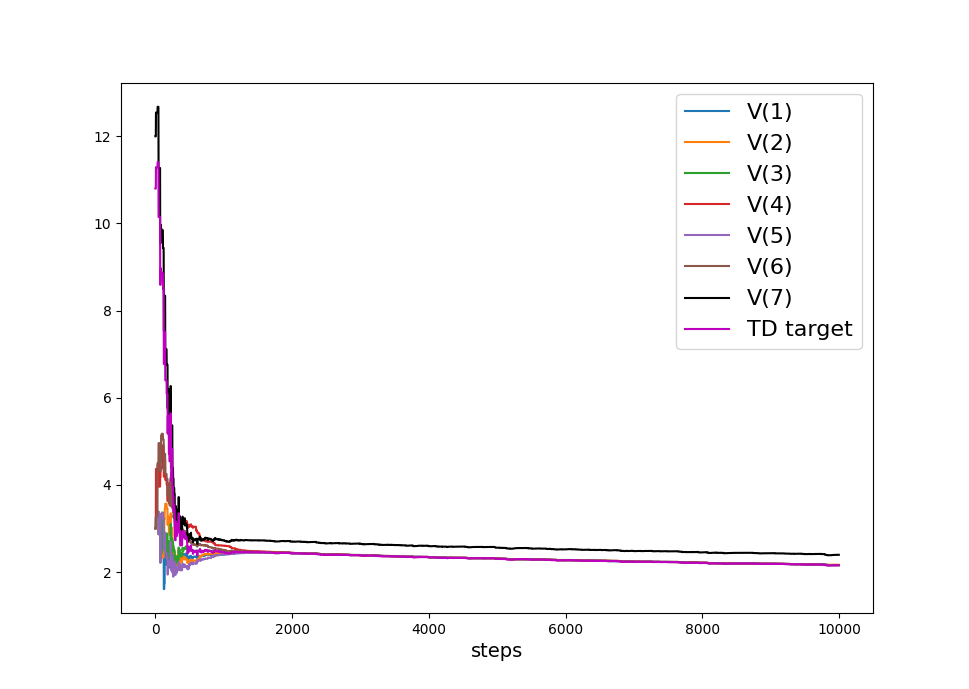}
\caption{Values of the seven states, and their shared TD target (discounted). This shows that for states 1 to 6, their values all match their TD target, while for state 7, it does not. }
\label{fig:baird_values}
\end{figure}

Figure \ref{fig:baird_values} shows all the approximated state values as well as their TD target which is shared, in this case just the discounted $V(7)$. This confirms the learning for the other six states didn't go wrong. The problem is that state $7$ did not provide a good learning target for them. This also shows that in general, not limited to this example, in practice even the TD error or the Bellman error is zero for some state, the value of the state is not necessarily correctly estimated. 

\begin{figure}[h]
\centering
\includegraphics[width=0.8\linewidth]{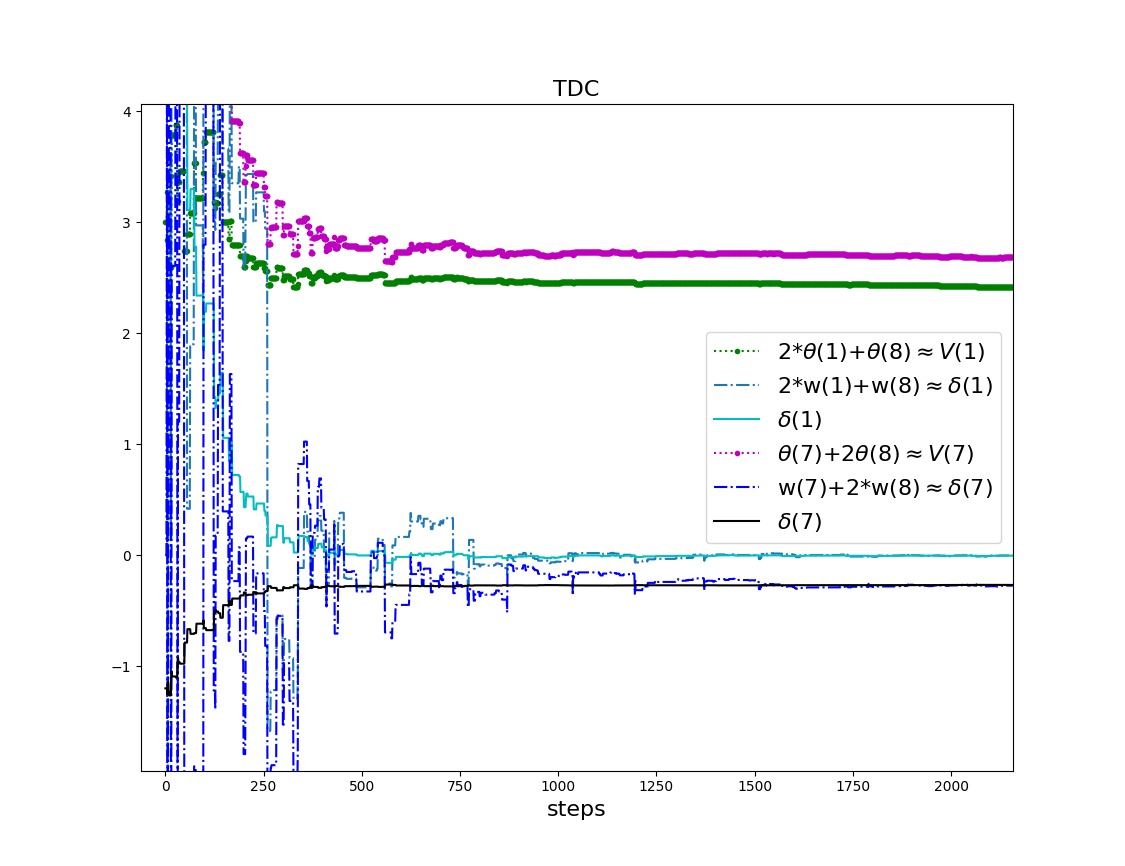}
\caption{Approximate values and TD errors of the seven states.}
\label{fig:baird_values_tderrs}
\end{figure}

\begin{figure}[t]
\centering
\includegraphics[width=0.8\linewidth]{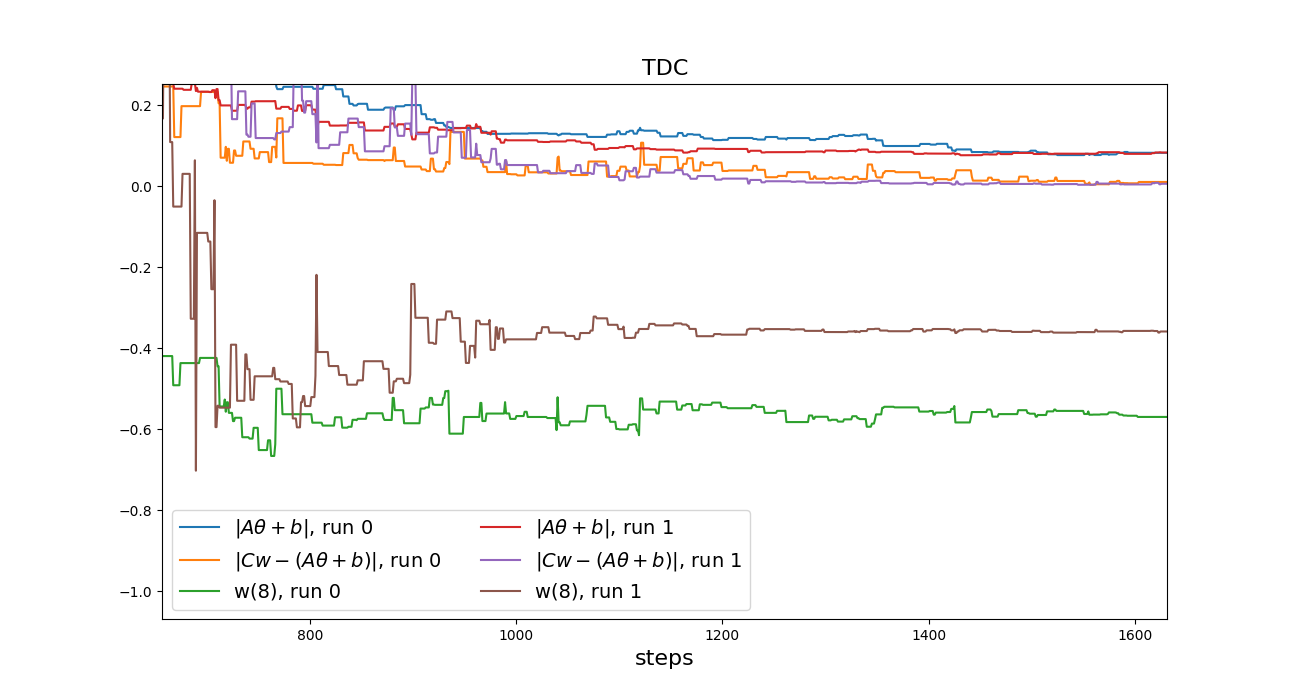}
\caption{Plots of the $\bold{ODE}(w|\theta)$ loss and the square root of NEU, in two random runs. The trajectories of $w(8)$ are also shown.}
\label{fig:ode_loss}
\end{figure}

We further break down our main lens, the regression error. Figure \ref{fig:baird_values_tderrs} shows the regression targets and predictions for the helper iterators. This corresponds to the plots of $\delta$ (TD error) and $\hat{\delta}(w)$ (the approximate TD error using weights in $w$). We focus on state 1 and state 7. For the other five state, it is similar to state 1. We also show the approximate values alongside the TD errors. 
We can see one problem of TDC in this figure: {\em The helper iterator has an early satisfaction that it shouldn't}. 
The TD errors are still high, but the helper iterator already predicts them so well! 
In particular, the TD errors of state 1 and up to state 6 converge to zero quickly. However, the TD error of state 7 is not yet. However, the helper iterator already does a good job in regressing the TD errors for all the states including state 7. It should not regress the TD errors so well before all the TD errors come close to zero. This is the key that lead to the slowness phenomenon in Figure \ref{fig:baird_tdc} and Figure \ref{fig:baird_tdc_alphas_betas}.

Now we are going to use our second lens:
\[
\bold{ODE}(w|\theta) = \|C w - (A\theta + b)\|.
\]
This is because we want to understand further the behavior of the helper iterator in solving the O.D.E., $\bold{ODE}(w|\theta)=0$. 
We plot this metrics during learning, as well as $\| A\theta+b\|$, which is just the square root of the NEU objective. 
The plots are shown in Figure \ref{fig:ode_loss}. The figure shows that, across runs, the solution of $w$ is different. However, all these solutions drive the $\bold{ODE}(w|\theta)$ to converge to zero in a short time. The NEU objective, instead, is not. This appears to be a learning bias, but it is actually not. In fact, TDC is finally ``jammed'', the main iterator is still moving in the right direction although the helper iterator almost already stopped. 
Overall, in TDC, the helper iterator regresses the TD errors well, but the main iterator does not reduce NEU effectively. Note that because the main iterator also uses the helper iterator for updates, we cannot say that it's only the main iterator's fault.

  Let's see why TDC is jammed. Let’s write the TDC update, in terms of the main iterator:
\begin{align}
\theta_{t+1} &= \theta_t + \alpha\delta_t\phi - \gamma (\phi^\tr w_t)\phi' \nonumber \\
&= \theta_t + \alpha\delta_t\phi - \gamma (\delta_t + \epsilon_t )\phi' \nonumber \\
&\approx \theta_t - \alpha\delta_t(\gamma \phi' - \phi). \label{eq:tdc_rg}
\end{align}
The first line is just the original TDC update for $\theta$. The second line expresses the prediction with the target plus an error term, $\epsilon_t$. The last line is because in the experimentation above (e.g., Figure \ref{fig:baird_w_regression}), we see that the TD error prediction is very accurate, and $\epsilon_t$ is small.  

Seasoned reinforcement learning researchers will immediately recognize equation \ref{eq:tdc_rg} is just the Residual Gradient (RG) algorithm by Baird, proposed in the same paper as his seminal counterexample \citep{baird1995residual}. So, surprisingly, {\em the main iterator in TDC becomes RG when TD errors are predicted well enough by the helper iterator}. This is a novel perspective of TDC that is not discovered before in literature, to the best of our knowledge. 

So the slowness and the extended flat learning curves in TDC is just caused by the slowness of RG. However, RG is not necessarily slow by itself. The catch is that in order to approach RG in TDC, we have to use small step-sizes. Otherwise, as we have seen in our experiments, oscillation will happen. A remedy for this is perhaps to increase the step-size $\alpha$ when the regression error by the help iterator becomes small. This observation may help us in developing step-size adaptation methods for TDC, which, however, is beyond the scope of this paper. Nonetheless, let's go back to the jammed behavior of TDC, finally boiled down to RG. Let’s study in particular the update for state 7, which is the bottleneck state. The transition is under the target policy, so for state 7, the update of equation \ref{eq:tdc_rg} re-writes into
\begin{align*}
\theta_{t+1}
&= \left(I - \alpha (1-\gamma)^2 \phi (7) \phi (7)^\tr\right) \theta_t.
\end{align*}
With the tuned, near optimal $\alpha$ equal to 0.005, one can show that this iteration contracts at a rate of $0.99975^t$!
Thus no wonder TDC in Figure \ref{fig:baird_tdc} and the best curves in Figure \ref{fig:baird_tdc_alphas_betas} are so flat. \citet{sutton2018reinforcement} did not make a mistake in interpreting their results. It is a phenomenon of slow learning, not a learning bias. 

\section{Afterthoughts}
So we conclude that TDC for Baird counterexample is a slow convergence problem, instead of a learning bias, improperly chosen step-sizes, or insufficient learning time. 
We also found that {\em in TDC, the helper iterator is satisfied too early, which almost perfectly regresses an TD error that is not perfect}. The afterthought for uncovering this phenomenon is, why did it happen?

Re-looking at Figure \ref{fig:ode_loss} resolves the myth. The plots show that, even though $w$ converges to different fixed-points across runs, the $\bold{ODE}$ loss, the underlying error for the helper iterator, drives to zero in each run. However, in the meanwhile, the NEU loss, does not go to zero in the runs. This is strange, because the $\bold{ODE}$ loss should be proportional to the NEU loss: 
\[
\bold{ODE}(w|\theta) = \|C w - (A\theta + b)\|; \quad \bold{NEU}(\theta)=\| A\theta+b\|^2.
\]
The only possibility is that matrix $C$ does not relay the $\bold{ODE}(w|\theta)$ loss to the NEU loss well. Surprisingly and sadly, this is just the source of the problem. 

Unnoticed by previous work, the matrix $C$ is a singular matrix for Baird’s example. This simple fact causes great trouble for TDC and GTD2, in a hideous way that we saw. In fact, $A$ is singular too. Note the feature vector is 8-dimensional and we have 7 states in Baird counterexample.  Both matrices $A$ and $C$ are 8 by 8. However, they are both formed by a product involving the feature matrix which is 7 by 8. Thus $C$ and $A$ are both singular. As a result, the MSPBE is not even well defined for the Baird’s example. A fix is to apply some perturbation to matrix $C$, or equivalently, $\ell_2$ regularization to the helper iterator. 
This is exactly what was done in the TDRC algorithm \citep{martha2020gradient}. However, the original authors did not observe this merit, because they motivated TDRC by mixing TDC and TD, to gain acceleration from TD. In their experimentation, they simply used the MSPBE metric, and the metric was computed with matrix pseudo-inverse.   In all, the slowness of TDC on Baird counterexample is due to the use of a singular preconditioner in MSPBE, because one can view MSPBE as a preconditioned NEU just like how TD and the Expected GTD are preconditioned \citep{ptd_yao}, with the hope of reducing the squared condition number of the normal equation in GTD \citep{impressionGTD}. \citet{ran_gtd} had similar opinions on this. Note, the   preconditioning effect of $C$ in MSPBE is realized in a {\em stochastic} way (this is very different from iterative algorithm in numerical analysis), because only samples of $C$ are applied in the learning update. Given this distinctiveness, the preconditioner is often not obvious to algorithm users in off-policy learning.  Thus users should pay special attention to the behavior of TDC and the use of MSPBE regarding this hidden trap. 

\section{Impression GTD on Baird Counterexample}
Recently, \citet{impressionGTD} proposed a new GTD algorithm, called Impression GTD. The algorithm features in a single-time-scale formulation of minimizing the NEU objective. Due to this formulation, essentially Impression GTD is the standard SGD algorithm, with only one step-size. It is much easier to use for practitioners than GTD, TDC and GTD2. The algorithm also has a mini-batch version. \citet{shangtong_imgtd} independently proposed a very similar algorithm, called Direct GTD, in the setting of infinite-horizon problem, with a convergence and convergence rate analysis under the diminishing step-size.

\begin{figure}[t]
\centering
\includegraphics[width=0.9\textwidth]{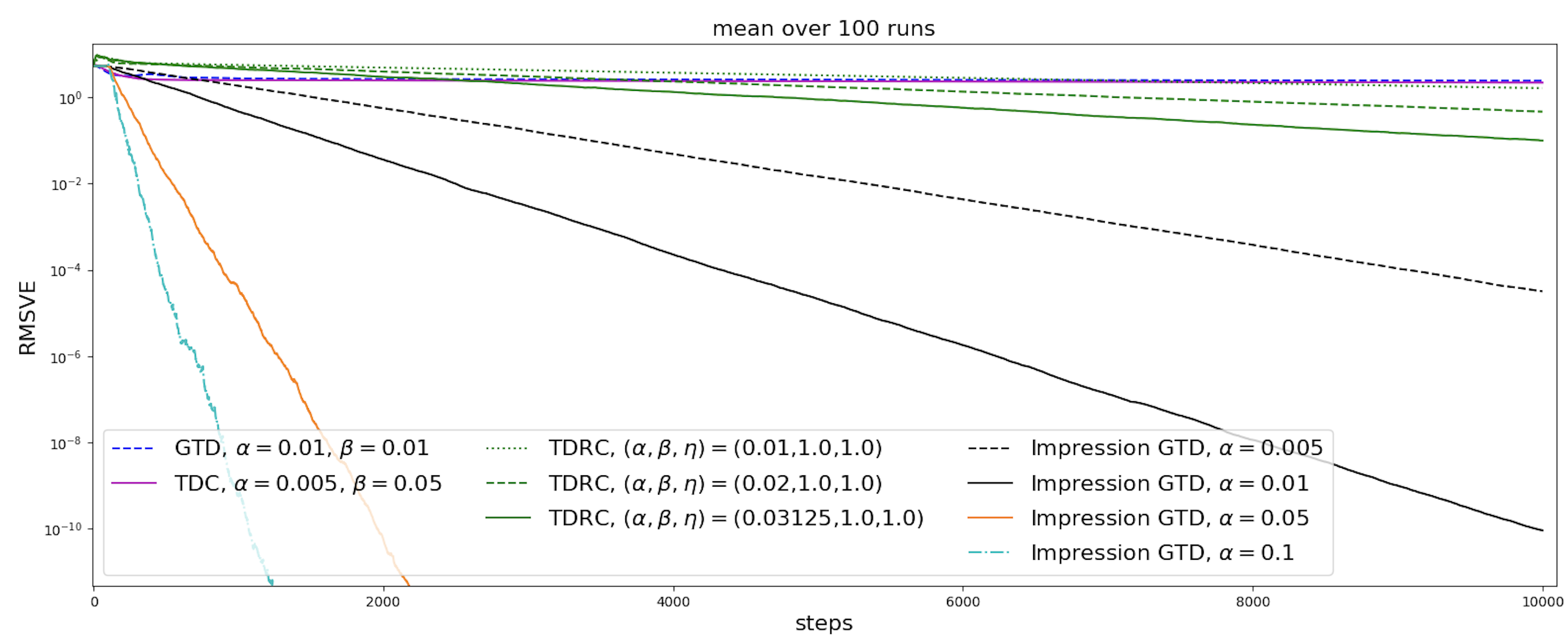}
\caption{Baird counterexample: Impression GTD versus GTD, TDC and TDRC.}\label{fig:baird_results}
\end{figure}

The performance of Impression GTD algorithm on Baird counterexample is shown in Figure \ref{fig:baird_results}. For a clear visualization, we only show the curves of GTD, TDC and TDRC. The other baselines were not as fast as the chosen baselines. 
For TDRC, all the hyper-parameters are used the same as the TDRC paper, which were selected by the original authors from an extensive sweep search. The $\alpha$ was 0.03125, $\beta$ (the regularization factor) was 1.0, and $\eta$ was 1.0, too. We also tried bigger values of $\alpha$ (without changing $\beta$ or $\eta$), including 0.04 and 0.05. They had either much bigger variances or diverged. Impression GTD used a batch size of 10. All the algorithms are corrected by $\rho$, the importance sampling ratio.   

The Impression GTD didn’t start learning until 100 steps of following the behavior policy, filling the buffer with some content. We see that $\ell_2$ regularizing the helper iterator is indeed helpful, and TDRC is much faster than TDC. Impression GTD agents learn even much faster, with a steep drop in the value estimation error, all the way down to near zero. With a small $\alpha$ like 0.001, the algorithm converges slower, but it also drives the RMSVE down to near zero. 

The curves of Impression GTD exhibit the pattern of a linear rate. Appendix A.6 of \citep{impressionGTD} 
also has a theoretical proof that Impression GTD converges linearly for Baird counterexample. 

\section{Conclusion}
This note contributes an in-depth debugging analysis of the TDC algorithm for off-policy learning on Baird counterexample, which is used as a standard test bed in reinforcement learning. The debugging process is well detailed, to be helpful for someone who work on two-time-scale stochastic approximation algorithms. 

We have some interesting observations discovered in the debugging process. To name a few, 
\begin{itemize}
\item TDC algorithm reduces to RG once the helper iterator does a good job for regressing the TD errors. 
\item In implementing multi-time-scale stochastic approximation algorithms, one has to make sure that the helper iterators are not satisfied too early. Failure to do so will incur the jamming behavior of the algorithm. Extremely slow convergence and a flat curve looks like a learning bias, but it is not. 
\item For debugging and analysis purpose, it helps to monitor the individual error for each iterator in the system. This can help us see which iterator(s) did well, and which went wrong. It then helps to examine the relay links between the broken ones. 

\item The conditioning of the problem has to be made sure beforehand. In the cases like MSPBE and TDC, one has to make sure solving a sub-problem in a helper iterator relays back to the main problem, and all the iteration errors go down simultaneously. An indicator of the relay being broken is that, the helper iterator does a good job in the error metrics that it's supposed to minimize, but the main learning objective error is still large.   

\end{itemize}
Finally, for Baird counterexample, with the guaranteed convergence and a linear convergence rate of Impression GTD, we safely conclude that the perplexing Baird counterexample is solved for off-policy learning. Note although RG also converges fast for this example, in general, RG is not guaranteed to converge to the TD solution, and the RG solution is often inferior, e.g., see \citep{tdc}. 
 
\bibliographystyle{plainnat}
\bibliography{sample}

\end{document}